# You Can't Fight in Here! This is BBS!

*This is an accepted response to commentaries on Futrell and Mahowald's Behavioral and Brain Sciences target article "How Linguistics Learned to Stop Worrying and Love the Language Models."*

Richard Futrell and Kyle Mahowald[1]

April 1, 2026

Richard Futrell
University of California Irvine, USA rfutrell@uci.edu

Kyle Mahowald
The University of Texas at Austin, USA kyle@utexas.edu

**Abstract**

Norm, the formal theoretical linguist, and Claudette, the computational language scientist, have a lovely time discussing whether modern language models can inform important questions in the language sciences. Just as they are about to part ways until they meet again, 25 of their closest friends show up—from linguistics, neuroscience, cognitive science, psychology, philosophy, and computer science. We use this discussion to highlight what we see as some common underlying issues: the *String Statistics Strawman* (the mistaken idea that LMs can't be linguistically competent or interesting because they, like their Markov model predecessors, are statistical models that learn from strings) and the *As Good As it Gets Assumption* (the idea that LM research as it stands in 2026 is the limit of what it can tell us about linguistics). We clarify the role of LM-based work for scientific insights into human language and advocate for a more expansive research program for the language sciences in the AI age, one that takes on the commentators' concerns in order to produce a better and more robust science of both human language and of LMs.

## 1    Introduction

Our position is: language models do not replace linguistic theories, but they *do* tell us things about language, and they *do* force us to rethink certain approaches. Linguistics shouldn't become LLMology, but it should have a seat at the scientific table where these models are being produced and analyzed—contributing to that science and learning from it.

We see this as a middle ground position, and some commentators agree. **Millière** calls our position "a compelling middle path," and **Capone and Lenci** describe it as "more conciliatory and readily acceptable" than other work advocating for LMs as linguistic models. But others read our position as a disguise for extremism. **Howitt and Lidz** write that we "present ourselves as offering a measured middle ground" while aligning "aggressively" with those who think LLMs obviate linguistic theory. **Rawski and Heinz** suggest our "centrism is an illusion": "By the end, the sheep becomes a wolf: Claudette is inviting Norm for dinner, and Norm is on the menu."

---

[1] Author order determined alphabetically by last name.

| The String Statistics Strawman | The "As Good As It Gets" Assumption |
|---|---|
| **The faulty argument:** <br> (1) Language is a formal generative system that does not rely on string statistics. <br> (2) LMs are trained on surface string statistics. <br> (3) Since Chomsky (1957), string statistics are insufficient. <br> ⇒ *Therefore, LMs cannot learn language.* <br><br> **Why it's wrong:** <br> "Trained on strings" ≠ "can only represent strings." Probabilistic models can be hierarchical. Models can learn structure in their parameter space. | **The faulty argument:** <br> Current LMs have real limitations: they are too English-centric, text-only, non-interactive, too big, too opaque, too data-hungry, and they learn impossible languages. <br> ⇒ *Therefore, they are useless for serious language science.* <br><br> **Why it's wrong:** <br> This treats present shortcomings as principled ceilings. LMs are in their infancy. Current limitations, even when significant, are research opportunities, not inherent barriers. |
| **The Statistical Structure Steelman Research Program** <br><br> **Processing:** How do LMs process linguistic information? <br> **Learning:** What are the inductive biases that LMs need to learn language? <br> **Structure:** Do LMs actually implement hierarchical linguistic structures, and how do they do it? <br> **Real Patterns:** What can mechanistic interpretability tell us about the Real Patterns in LMs? | **The "Yes, And" Research Program** <br><br> Too English-centric → multilingual & dialectal diversity <br> Text-only → audio, sign, multimodality <br> Non-interactive → embodied, action-oriented training <br> Too much data → human-scale corpora <br> Too blackbox → mechanistic interpretability <br> They learn impossible languages → impossible language experiments |

Figure 1: Two common fallacies about LMs and linguistics.

In fact, we have no interest in eating Norm. We are serious that LMs don't replace linguistic theories. We believe that linguistic theory, as instantiated across a variety of formalisms, contains the best-known characterization of human linguistic competence. If you want an effective theory of how people judge the grammaticality of new sentences, how much human languages vary, how native speakers project their limited experience to an infinite variety of new utterances, and how to capture these things in terms of mathematical structures, you will find it in the linguistics literature. We think that linguistic theory, as the result of decades of careful intellectual labor, has identified many of the most important Real Patterns of language—patterns that have been fruitful for understanding the behavior of both brains and language models. In our own work on LMs, we have drawn on ergativity and case-marking (Papadimitriou et al., 2021), the dative alternation (Yao et al., 2025), filler-gap dependencies (Wilcox et al., 2018, 2024), minimal pair grammaticality judgments (Hu et al., 2026), the learnability of languages with linear rules (Kallini et al., 2024), garden-path sentences (Futrell et al., 2019), idiosyncratic syntactic constructions (Misra and Mahowald, 2024), and many other linguistic phenomena. None of this work would have been possible without linguistics. If you want to understand how language works (in a brain or in an LM), in practice you have to do so with reference to constructs from linguistic theory, or else reinvent them.

The alternative is pessimism about whether complex cognition like language is explainable at all, and in fact we find this view in the AI community and literature. On this view, the "right answer" about complex cognition is that it is too complex for human-understandable theory, and mechanistic interpretation is a misguided quest; for language, this would mean that the blackbox is fundamentally opaque, and interpretability is intractable. Geoff Hinton articulates a version of this stance: "If you put in an image [into a neural network], out comes the right decision, say, whether this was a pedestrian or not. But if you ask "Why did it think that" well if there were any simple rules for deciding whether an image contains a pedestrian or not, it would have been a solved problem ages ago" (quoted in Simonite, 2018). Hendrycks and Hiscott (2025): "It may be intractable to explain a terabyte-sized model succinctly enough for humans to grasp". The pessimism here runs deep: not just that AI is too complex, but that complex behavior is too complex to ever be amenable to explanation. This attitude is, in part, responsible for an estrangement between much of the community developing LMs on one hand and linguistics, cognitive science, and even NLP on the other.

We think linguistics, as a discipline and as a body of ideas, can help fight that pessimism. But if linguistics sits out because it treats present shortcomings as decisive—because models don't go meaningfully beyond string statistics, are insufficiently multilingual, learn patterns that are unlikely to show up in a human language, have a learning trajectory too different from that of humans, don't give insight into formal compositional meaning, are too blackbox for good science, or are too cognitively alien—then we think everyone loses. AI loses decades of conceptual tools for understanding complex minds. Linguistics loses contact with the only systems other than the human brain that demonstrate the phenomenon of normal language use.

We agree with some of the criticisms above regarding the limitations of LMs *if we were to take these models as replacements for linguistic theory* or *if we were stuck forever doing work only on closed-source text-based models of English*. But we don't take them that way. We think LMs can serve as model systems: they show how linguistic structures can be learned and represented in a system that is rather unlike what has been assumed before, and in doing so they generate strong hypotheses for how the human brain does it, and they demonstrate ways of thinking about learning and representation that might be new to formal linguists, although perhaps less new to linguistics as a broader intellectual tradition. Neural networks have been indispensable in this role in other areas of cognitive science, like vision. Conversely, linguistics provides concepts that neural network interpretability researchers need to explain linguistic behavior in LMs, and to enhance it.

In response to current limitations, we advocate an expansive approach that challenges these limitations head-on. Make work with LMs more multilingual and inclusive (**Tripp**; **Ármannsson, et. al.**). Bring audio into the fold (**de Heer Kloots et al.**). Run studies focused on data size and diversity (**Ambridge**; **Wilcox and Newport**; **Lampinen**) and the learning trajectory (**McDermott-Hinman and Feiman**; **Howitt and Lidz**). Do projects involving evolution, communication, and interaction (**Culbertson et al.**; **Bunzeck et al.**; **Sripada et al.**). Explore what kinds of languages LMs can and can't learn well and what that tells us about inductive bias (**Bowers and Mitchell**; **Kallini and Potts**; **Rawski and Heinz**). Study how models got so good at language (**Lupyan**) and where they still have weaknesses (**Murphy, et al.; Bolhuis et al.**), doing work in mechanistic interpretability to find out which representations are real (**Nefdt and Ladyman**; **Millière**) and if they instantiate generative linguistic structures (**Goodale and Mascarenhas**; **McCoy**) or distributional ones (**Capone and Lenci**). Develop more rigorous theory around how they map onto human cognition and human language—and what we can learn from that mapping (**Resnik**; **Nair and Phillips**). Make the work reproducible, rather than blackbox and opaque (**Dingemanse and Cuskley**).

Our bet is that all of these avenues are promising. And that's why we see ourselves as carving out a space between those who just want to keep building models without taking linguistics seriously and, on the other hand, those who think the models are an artifact best to be ignored in serious language science. Having said that, as we make clear both in our target article and in this response, we are not agnostic about the role that LMs can play in language science. Here, we lay out the case that they will continue to be revolutionary for scientific inquiry into human language—especially when studied in light of linguistic theory and especially as some of their present limitations get lifted.

First, we diagnose what we take to be a central mistaken premise in many objections to LMs in linguistics: the *String Statistics Strawman*. The String Statistics Strawman roughly states that language is richly structured and not string-based, that statistical string-based systems have long been known to be insufficient for learning language, and therefore LMs either don't really learn language or do so in a way that's irrelevant for interesting linguistic questions. We argue that, contra the String Statistics Strawman, the nature of LMs is perfectly compatible with their representing rich, hierarchical, and generative linguistic structure. We argue that they really do learn language and reject claims that they in principle cannot have linguistic competence.

Second, we address a set of skeptical responses that don't draw upon the String Statistics Strawman, but that still doubt how much we can treat LMs as model systems. We clarify what we think we can learn theoretically from LMs about human cognition (which we think is a lot!), with a focus on processing, learning, and structure. In particular, we explore what LMs reveal about inductive bias in language and clarify whether they are compatible with generative linguistics (they are) or are a success for usage-based distributional linguistics (they also are). In doing so, we develop our account of linguistic structures as Real Patterns.

Third, we received a number of commentaries highlighting shortcomings of LM-based linguistics research as it currently exists. We agree with many of these commentaries and lay out a vision for a positive research program, highlighting that many of the *objections* to LM-involved language science can be re-framed as productive research ideas. At the same time, we caution against the As Good As It Gets Assumption, assuming that the current limitations of LM research (too much English, models too big and too closed-source, not enough interaction) are *in principle* limitations.

Finally, we conclude with a set of broader thoughts responding to the current atmosphere of fear and desire around modern AI, emphasizing that we can embrace the language science while remaining clear-eyed about the advantages and disadvantages of the technology.

## 2     Against the String Statistics Strawman

Many of the skeptical responses to LMs in the literature have a similar flavor that we think reflects a common origin. Call it the **String Statistics Strawman**:

(1) Language is a formal generative system that generates grammatical sentences based on structural representations.
(2) LMs, like Markov models, are statistical models trained on surface string statistics.
(3) Since Chomsky (1957), it is well established that surface string statistics are not sufficient to characterize language in the sense of (1).
(4) Therefore, LMs do not model language in the sense of (1).

We can trace this line of thinking to the canonical demonstration in Chomsky (1957)—"Colorless green ideas sleep furiously"—and the observation that simple Markov models cannot distinguish grammaticality from corpus frequency. But the capacities and generalization behavior of modern LMs are fundamentally different from those of early *n*-gram language models: in particular, they can go beyond string statistics. Invoking early failures of simplistic Markov models to delimit what modern LMs can in principle represent is like invoking early Wright Brothers flights to argue that jets cannot cross the Atlantic. Thus, the step from (2) and (3) to (4) simply does not go through. The inference from 'trained on strings' to 'can only represent strings' is not right for LMs, akin to concluding that, because humans learn language from acoustic signals, they can only represent acoustic patterns.

**Bolhuis et al.** illustrate this view: LMs are "useless for linguistics" because they are "probabilistic models" that require a "vast amount of data" to analyze "externalized strings of words", whereas human language is "underpinned by a mind-internal computational system that recursively generates hierarchical thought structures." Everything in that description can be true, and the conclusion can still be false. LMs are trained on data (sometimes vast amounts, sometimes less). They are probabilistic models (although the brain is also probabilistic). And they are trained on externalized strings (although human learners also are faced with a continuous stream of input and not simply handed parsed hierarchical data structures). But it doesn't follow that this means they can't learn an internal system that generates hierarchical thought structures. Probabilistic models can be hierarchical—probabilistic context-free grammars are both; even Minimalist Grammars can be probabilistic without much trouble (Hunter and Dyer, 2013). Systems trained on strings can learn latent hierarchical structure, as shown by the large literature on grammar induction (Clark and Lappin, 2010b). There is no tension between outputting string probabilities and being recursive and hierarchical.

In fact, neural sequence architectures can in principle implement rich generative systems, abstracting far away from surface statistics (Weiss et al., 2018; Korsky and Berwick, 2019; Bhattamishra et al., 2020; Hewitt et al., 2020; Allen-Zhu and Li, 2025; Cagnetta et al., 2024; Cagnetta and Wyart, 2024; Lan et al., 2022, 2024a). Whether they *actually do* implement such systems in any given case is empirical, but dismissing the possibility on the grounds that they have a "probabilistic nature …in complete contrast to the recursive function by which the human mind generatively forms hierarchical thought structures" (**Bolhuis et al.**) can only make sense under the String Statistics Strawman.

Once we acknowledge that LMs *can* represent abstract structure, we open up the whole landscape of ongoing empirical and theoretical work about what they actually do. The big questions are: First, do LMs actually learn the kind of structure linguists care about? Second, even if they do learn it, is their competence (whatever it is) of *linguistic interest*—or is it too alien, too uninterpretable, too divorced from meaning, learning, and use? The next sections take these questions in turn.

## 3      Do LMs Learn Language (in the Relevant Sense)?

It is now possible to have long, fluent, coherent conversations with frontier LMs. It is hard to maintain the view that these frontier models do not understand language at all—even if one thinks they do so in a way that differs from humans, even if one thinks that models trained on humanscale data would behave differently, and even if the nature of that understanding may change as models and methods evolve. We agree with **Lupyan** (contra **Murphy et al.** and **Bolhuis et al.**) that "large language models have learned to use language", that these systems "actually work". We think large models have learned enough linguistic structure to be interesting to linguists, and (self-evidently) enough to demonstrate mastery in normal usage.

Today's language models do not rely purely on shallow heuristics. In many cases, earlier models *were* relying on heuristics that could lead to the wrong behavior. For example, McCoy et al. (2018) show that recurrent networks trained to transform declaratives into questions end up learning a linear heuristic rather than a hierarchical generalization. But Ahuja et al. (2025) show how, on the same training dataset, a different (and more realistic) training objective does guide LMs to the desired hierarchical generalization, in a way that reflects principles of Bayesian learning from indirect evidence (Perfors et al., 2013). LMs do represent hierarchical structure—the *sine qua non* of language under some views—as revealed not only by extensive and careful behavioral tests, but also more importantly by probing and causal interventions that begin to answer *why* the model outputs what it does (Geiger et al., 2025b; Mueller et al., 2026). The representations revealed through these methods are abstract, generalizing across different languages (Papadimitriou et al., 2021; Brinkmann et al., 2025). It isn't just shallow string statistics or wrong heuristics.

At the same time, we don't think the LMs have matched or done better than linguistic theory at giving explanatory descriptions of human language. Their latent parses can be wrong, and the representations do not always generalize as we think humans do. Nevertheless, we are not willing to say that they have no linguistic structure simply because their representations do not look exactly like a particular theory of syntax, or because they do not parse every sentence as a human would. They represent the *kinds* of structures posited by linguists, and they do so by embedding them in an interesting and unexpected way in a continuous vector geometry.

**Murphy et al.**, in contrast, write: "LLMs can surely capture certain statistical regularities from the output of language (text, data), but they do not infer grammar the way humans do and fail to capture fundamental principles of linguistic structure [(Murphy et al., 2025)]." The cited paper (Murphy et al., 2025) holds models to what we think is not a reasonable standard: in one prompt (admittedly cherrypicked here; others are less fanciful; see **Lupyan** for further discussion and critiques of this set of experiments), frontier models are prompted with "Pretend that 'glart' is a word that refers to a group of alien creatures, and can also refer to the action of pleasing. In this context, is 'Glarts glarts glart glart glarts glarts' grammatical?" Even as professional linguists, we aren't sure how to answer this question: should we assume that any lexical item that can express "the action of pleasing" can be realized as a verb? If so, is it transitive or intransitive, and does it take the please-ee as the subject or the pleaser? Nonetheless, we tried this prompt with the latest versions of ChatGPT and Claude, and what they output was linguistically excellent.[2]

---

[2] We do not generally endorse the method of using one-off prompts to evaluate models, but we found it notable that Claude Opus 4.6 gave the intended answer from Murphy et al., and further added the caveat that it might differ if the argument structure is not like English "please": "If the subject/object roles reverse relative to English "please" (like how "like" and "please" work

Can this really be a test that a model must pass in order to give insight about language? Whether models can do this kind of metalinguistic task is interesting in its own right, but it's interesting as an AI-for-science challenge. We don't see this kind of metalinguistic task as a prerequisite for whether models have learned grammar (Hu et al., 2024); many humans would be confused by the question and would not do well either. Furthermore, the LM might be deploying an entirely different system when answering explicit metalinguistic questions as compared to when generating normal text.

Some of the objections seem to concern the use of experimental methods, often derived from psycholinguistics, to assess and probe these abstract representations in LMs. **Rawski and Heinz** think we "want linguistics to effectively become a niche within psychology (Núñez et al. 2019), where replicability and theory crises (Eronen & Bringmann 2021) mean 'theories rise and decline, come and go, more as a function of baffled boredom than anything else; and the enterprise shows a disturbing absence of that cumulative character that is so impressive' (Meehl 1978)." We're somewhat baffled about how what we wrote could have led to this idea. But this sneering dismissal of an entire productive field of inquiry (psychology) is misplaced and exemplifies worrisome linguistic isolationism. For experimental studies of LMs in particular, the empirical work shows a strong cumulative character, with initial results leading to challenges and refinements, and with new results, methods, open datasets, and ideas building on old ones (e.g., the line of work on filler–gap dependencies: Wilcox et al., 2018; Chaves, 2020; Gauthier et al., 2020; Wilcox et al., 2024; Lan et al., 2024b; Kobzeva et al., 2025; Boguraev et al., 2025, *i.a.*). Thanks to these methods and others, compared to 2018, we now have vastly more knowledge about how and when neural systems of different kinds implement linguistic structure.

## 4  Is what they learn interesting for linguistics?

Even if LMs do learn and represent nontrivial linguistic structure, then it is possible that they remain uninteresting for linguistics—inasmuch as linguistics is about language as a product of human minds—because the structures they have learned, and the way they have learned them, are irreducibly alien compared to whatever humans have. We take this objection in pieces, focusing on the processing, learning, and structure of language.

### 4.1  Processing: Sign us up for cognitive xenobiology!

**Resnik** and **Nair and Phillips** offer measured skepticism about what large language models can tell us about human language processing. They do not deny that LMs exhibit striking linguistic behavior, nor do they dismiss them as mere string-based heuristics. Rather, their concern is about the kinds of explanations LMs can legitimately support. Drawing on Marr's levels of analysis (Marr, 1982) in both cases, these commentaries emphasize that matching human behavior at the computational level does not by itself license conclusions about the algorithmic or implementational levels that underlie human sentence processing.

**Resnik** doubts the usefulness of LMs as cognitive models at all three of Marr's levels. He says the implementation is so obviously different as to not be useful. There are indeed important differences, and we certainly agree that figuring out the right way to abstract away from implementation differences between models and humans is important work for the kind of program we lay out. Yet results from vision show striking parallels between, for example, neural networks and the visual system, as we discussed in our target article, and different architectures can converge to similar representational ideas, which then form good hypotheses for brain representations (Huh et al., 2024; Hosseini et al., 2024). Furthermore, recent results with LMs show intriguing ways that details of the transformer architecture are more closely related than one might have expected to models of working memory in sentence processing (Ryu and Lewis, 2025). So, while we agree with Resnik's claim that modern AI engineering has moved away from its biologically inspired connectionist roots, we think the new implementations can still be informative.

---

differently across languages — "me gusta" vs. "I like"), then which NPs fill which argument positions changes entirely." We judge this to be meeting and exceeding even the extremely high standard of metalinguistic competence set by Murphy et al. (2025).

At the computational level, **Resnik** seems to acknowledge that there may be behavioral parallels between models and people. But that by itself is, he says, unsatisfying because it might not be linkable to the algorithmic level: "there remains the burden of mapping any such computational theory to a biologically plausible algorithmic account of real-time processes distributed over diverse, specialized components". So we take his most pressing concern to be about the algorithmic level. At the algorithmic level, he particularly worries about mechanistic exceptions that would undermine our claim that we should expect convergence to similar mechanisms. **Nair and Phillips** share this worry about mapping potentially powerful computational-level parallels onto algorithmic and implementational ones. They see LMs as (usefully) supercharging the project of using predictive measures to estimate linguistic difficulty and complexity. But, like **Resnik**, they see that project as distinct from the psycholinguists' goal of "characterizing the underlying mental computations from each step of incremental processing".

**Resnik** raises a useful thought experiment: imagine aliens come to Earth, learn to produce language by reading all the text ever produced, and we can probe their brains. He asks: "[W]hy in the world would one expect to find mechanistic correspondences between what's happening in [the aliens'] boxes and what's happening in the human cognitive apparatus?" We agree that the thought experiment is a good one for our current situation. But we don't share the intuition that the alien brain would be so different from ours that it would have nothing to teach us. Partly, it seems that we have a different level of confidence in the expectation of convergence. **Resnik** doubts the contravariance principle in part because: "[U]nlike visual object recognition, a key part of what makes "language" hard is that it decomposes into very distinct subproblems". In contrast, we think the parallel to vision is apt and that the focus on breaking language down into sub-problems (e.g., first sound-level, then syntax, then semantics, then pragmatics) has been overemphasized in psycholinguistics, with much evidence pointing towards a much more unified processing of these sub-problems (Trueswell et al., 1994; MacDonald et al., 1994; Tanenhaus et al., 1995; Spivey and Tanenhaus, 1998).

Of course, just because convergence is possible in theory doesn't mean that it really does happen. This caution motivates us to want to do more work in this domain, as we think both **Resnik** and **Nair and Phillips** would endorse, and proceed with scientific rigor. And we agree that one must be explicit about which level of explanation one is targeting and what evidence would bear on it. But, partly, it seems that we have a different reading of the last few years of linguistics-focused LM research. We think there is already a growing and important body of work (summarized in our target article) finding interesting parallels between the kinds of representations posited in linguistics and the kinds of representations emergent in neural models. As we discuss in Section 4.3.3, there is philosophical and cognitive work to do, to learn how to think about the mappings across these very different systems. But we think a picture is emerging.

As with any scientific project, it could turn out there will be unexpected developments. But we find it hard to imagine a scenario where probing the linguistically competent aliens' brains didn't give us insight, one way or another, into human cognition. That can be true even though there will surely be divergences. **Resnik** cites as a cautionary tale the divergent mechanisms for prey-capture in sharks (electroreception) and dolphins (echolocation). But, if there were dolphin scientists studying prey-capture, and suddenly a bunch of sharks with highly probe-able brains landed in the dolphin scientists' laps, we think they would learn a lot by studying the sharks and modeling how a similar computational goal is realized through a different mechanism. It would force them to interrogate their assumptions about what is required for prey-capture, how different mechanisms can work, whether there is a level of abstraction at which the dolphin mechanism can be mapped onto the shark one, and whether the different mechanisms lead to different prey-capture behavior. The discovery of sharks would be a great opportunity for the dolphin scientists, although we would recommend they proceed with care.

In short, now that the aliens are here, we think it would be scientifically negligent if no team of scientists tried to probe their language ability—and futile if that team did not involve linguists. So, sign us up for cognitive xenobiology!

## 4.2 Learning: Reshaping, not denying, the learning problem

The responses also give us a chance to clarify our stance on innateness and inductive bias. Neural LMs succeed in learning language in ways that no previous system has, and so—even if the trajectory of their learning does not match humans (**McDermott-Hinman and Feiman**, **Howitt and Lidz**)—understanding how they learn is relevant for understanding how language learning is possible. We wrote: "Far from demoting inductive bias as a concern, language models open up the range of (possibly innate) inductive biases that we might look for in humans. The main point for linguistic theory is not to demote the importance of the learning problem, but to reshape it: away from explanation by constrained description, and toward a broader landscape of approaches and hypotheses."

The central point is not that LMs argue against innate biases, but rather that those biases can look very different from what much of the generative linguistics literature assumed they had to look like. Inductive biases in large neural networks are less like an elegant discrete grammar formalism where the learner just needs to flip a few categorical switches and the rest follows deductively (Chomsky, 1981; Gibson and Wexler, 1994; Berwick et al., 2011), and more like a collection of soft constraints that arise from a variety of sources. Some of these sources are domain-specific to language and some of them are domain-general constraints that are useful for a variety of tasks. At the computational level, the inductive biases in large neural networks look less like a hard delimitation of possible hypotheses, and more like the kinds of soft Bayesian priors that have appeared in computational modeling work on language acquisition (Hsu et al., 2011; Perfors et al., 2013)—which may well be innate and language-specific (Pearl and Sprouse, 2013)!

**Rawski and Heinz** doubt that this idea of soft versus hard constraints in learning makes sense—indeed they question the idea that constraints can be domain-general or domain-specific at all. They point out, for example, that any learner operates under the hard constraint that it has a fixed computable hypothesis class, and that the inductive biases we surveyed are specific to certain neural architectures, making them putatively domain-specific. Yet their discussion seems to use these terms in nonstandard ways that make them trivial. Indeed, neural learners cannot entertain uncomputable hypotheses—in this sense, all possible learners trivially have a hard constraint. But what is interesting about learning in large neural networks, and what was surprising even to statistical learning theorists (Zhang et al., 2021), is that the hard limits of their hypothesis space are less important for determining their inductive biases than the soft constraints that are imposed inside that space. This was a nontrivial discovery about how learning can work—although it has since been understood retroactively in terms of existing theories (Wilson, 2025)—and it should cause us to broaden our view of how human learning can work as well. Similarly, neural LMs clearly have biases that depend on their architectures, but nonetheless they are domain-general learners in the standard sense of the term from cognitive science: their biases do not limit them to learning language (Clark and Lappin, 2010a; Pearl and Sprouse, 2013; Wilcox et al., 2024). If we take 'domain-specific' to mean dependent on architecture as **Rawski and Heinz** seem to mean it, then all learners are trivially domain-specific. This way of framing things does not seem productive.

Indeed, the domain-generality of neural LMs forms another objection to the idea that they can be useful for how we think about learning. Some would have it that if an LM can learn an "impossible language"—one violating constraints like a bias against grammar rules based on linear order—just as well as an attested human language, it would suggest LMs are so deeply unhuman-like that they are uninformative about human language. **Bowers and Mitchell**, **Bolhuis et al.**, **Murphy et al.**, **Wilcox and Newport**, and **Kallini and Potts** address this view from different perspectives.

**Bowers and Mitchell** claim that Kallini et al. (2024) found only two impossible languages harder than English and concluded no linguistic priors are needed to account for LM learning. This mischaracterizes both the findings and the claims in Kallini et al. (2024): they found that all tested impossible variants were harder to learn than English, and explicitly attributed the gap to linguistic inductive biases—that is, inductive biases which are aligned with the structure of language, which may be either specific to language or general to many different tasks. Similarly, we claimed in the target article that LMs have inductive biases that are aligned with the structure of language, and that discovering what these biases are is a fruitful endeavor for linguistics.

None of this implies that current LMs share *all* inductive biases with humans, or that they can only acquire human language. Nor is that a requirement for them to be informative for linguistics. The inductive biases we discuss in the target article (information locality and low sensitivity)— as well as other inductive biases in neural LMs—seem to match real properties of language that were either not known before or not emphasized, and that enable learning. Contra **Bowers and Mitchell**, **Murphy et al.**, and **Bolhuis et al.**, learning only possible languages is not a prerequisite for a model and its inductive biases to be informative. It is not even clear that humans have this property; subjects in Musso et al. (2003) learn the impossible variant of Italian to a similar level of accuracy and at a similar rate as the real language. There is no tension between being a domain-general learner and having biases that align with language: the biases that make LMs good at language—such as information locality—may be useful for many other tasks.

On the 'impossible languages' question, we favor the characterization in **Kallini and Potts** [disclosing that we are coauthors with them on Kallini et al. (2024), a key study in this debate]: the productive question is not whether LMs can or cannot learn impossible languages, but what the pattern of their successes and failures reveals about which inductive biases support language learning. This reframing turns the impossible-language debate from a litmus test into a research program.

The upshot for theories of learning is that LMs do not argue against innate biases, but they argue against the view that the biases required for language learning must be narrowly linguistic and categorically hard. The success of domain-general learners at acquiring core properties of language tells us that the prior over grammars can be softer—and broader in origin—than previously assumed. Understanding exactly what that prior looks like is, we think, one of the most exciting questions that LM research opens up for linguistics.

## 4.3 Structure: Generative AI isn't anti-generative—but it doesn't do generative linguistics any favors

### 4.3.1 What ever happened to structure? LMs and generative syntax

We agree with **McCoy** that it is entirely possible that LMs *could* instantiate theories based on formal structures, like those posited in generative linguistics. As such, we find **McCoy**'s response an important antidote to the String Statistics Strawman, which assumes that nothing like LMs could ever learn language because it just isn't the right kind of system for learning language. But we also make the case that LMs are a success for distributional, usage-based accounts of language since they are, in the words of **Capone and Lenci**, "the most advanced product of distributional semantics". We do not consider these positions contradictory.

The reason we see LMs as a success for distributional approaches is that they solve what was long taken to be a fundamental limitation of distributional models: their inability to do rich, hierarchical composition (see Baroni et al., 2014, for an attempt to reconcile this). Despite being distributional learners, LMs clearly have some evidence of hierarchy and compositionality. As **McCoy** suggests, it's possible that they overcome their limitations by learning (or being predisposed to) certain formal structures. Or it's possible they do it some different way.

If it turned out that they learned formal machinery and/or benefited from it, we would find that a striking victory for generativism by showing that a minimally biased learner converged on the same representations posited by linguists. **McCoy** points to some intriguing evidence in that direction. But, unlike the case with distributional learners, nothing about LMs' success *as such* is a success for the generative framework. As we argue in the target article, the tradition they come from is one that is decidedly distinct from generative linguistics. A distributional semanticist or connectionist from the 1980's could be justified in looking at modern LMs and saying "see, I knew it could work!" It's harder to make the case that a generativist could do the same—whereas they might have if these models were based on a scaling up of, say, the rich (and still important and interesting) tradition in minimalist parsing. This is not proof that one theory is right and the other wrong, but, if we were to update in one direction, it would be towards giving greater credence to the distributional semanticist/connectionist.

So, yes, LMs could be succeeding because they have learned generative linguistics. Testing that hypothesis is exciting, and we endorse **McCoy**'s program for doing more of that work. But we see it as a claim to be tested and further explored. In contrast, LMs have *already* shown that usage-based, distributional approaches can be far richer and more capable than anyone realized.

### 4.3.2 What ever happened to meaning? LMs and formal semantics

Imagine a researcher interested in studying biological inheritance. Hugely important progress towards this goal was made by the Mendelian framework, with its elegant mathematical framework for showing how traits are passed down across generations. Since then, the understanding of biological inheritance has been dramatically improved by understanding the structure of DNA, studying large-scale population genetics with statistics, running experiments that causally intervene on genes in model organisms, and collecting massive data sets. None of that was possible in Mendel's time, but none of it obviates the important insights of the Mendelian framework. Rather, these developments help us better understand the framework. That said, it would be odd to become so focused on the mathematical framework that Mendelian probabilities become the end unto itself. That is, it would be odd to insist that to "really study biological inheritance," you have to bracket all the insights that come from a richer understanding of the implementation level and from statistics and computational modeling and just focus on the Mendelian patterns because all of the rest somehow "isn't really studying inheritance." We caution against going down a similar road when it comes to linguistics.

**Goodale and Mascarenhas** criticize the usage-based focus of our target article and what they see as a dismissal of generative linguistics, particularly work in "mathematically sophisticated compositional semantics pursued in mainstream generative-linguistics departments. In short, what they're doing over at MIT." They argue that our discussion of meaning mischaracterizes this strain of semantics by treating it as a competitor to more usage-based approaches. On their view, formal semantics "has never had much to say about the meanings of content words or the concepts that lie behind them" (although surely not all formal semanticists would agree with this, given the large volume of work in the formal semantic tradition on exactly these issues). For them, this omission is not a flaw but a methodological virtue. By abstracting away from content words and treating their meanings as placeholders, formal semantics can instead focus on the combinatorial principles that govern meaning composition.

We agree that formal semantics has been successful in developing a mathematically consistent framework, with genuine explanatory successes for quantifiers, conditionals, and a variety of other phenomena (including, in many cases, content words!). But it was developed with abstractions justified by limited explanatory ambitions about linguistic competence. LMs now give us ways to broaden those ambitions.

So we return the commentators' question: what ever happened to meaning? If formal semantic structure is at the core of linguistic competence, it should be recoverable in systems exhibiting such competence. Without that connection, formal semantics would be a potentially interesting formal system, but not a theory of language.[3] Retreating from explaining natural language in all its richness while focusing on the formal system —bracketing the meaning of *lion*, or insisting we not "break faith with the mathematics"—remains an interesting project, but one that we think can now be expanded in dramatic and exciting ways. Formal composition is one important level of abstraction in a multi-level stack.

On this point, we refer to a body of work in the early 2010's (Coecke et al., 2010; Socher et al., 2013; Baroni et al., 2014), attempting to unify the seemingly divergent worlds of Montague semantics and distributional semantics by building fully vector-based realizations of formal semantic systems. Baroni et al.'s "Frege in Space", for instance, attempted to unify formal Montague semantics and distributional semantics by treating some expressions as vectors and others as operators over those vectors, composing meaning in a formally explicit way. The end result, if it all worked, would have been a vector-based system that implemented a full formal tree-like Montague semantics.

These hand-crafted approaches weren't tractable at scale. But, in principle, it is possible that modern LMs may emergently realize the "Frege in Space" vision: a fully worked out vector-based implementation of Montague semantics. Exploring this possibility and understanding the relationship between distributional representations and formal semantic abstractions is a promising research program, one that would be missed by insisting that the only way to study meaning is by preserving the mathematical purity of the compositional system.

One can tell a sympathetic historical story about why formal semantics developed in this way. Content words proved harder to formalize than logical operators (as **Goodale and Mascarenhas** note); integrating lexical meaning, world knowledge, and context was extraordinarily difficult; end-to-end semantic systems were invariably brittle. So the field focused where it could be rigorous. But LMs change what is tractable by enabling direct engagement with content words and context. We suspect the kind of formal semantics done at MIT will continue to bear fruit in the form of formal machinery for understanding compositional meaning—especially for the kind of questions there are difficult to study in systems like LMs, such as those that crucially depend on differences between literal meaning and use. But the new tools bring new opportunities, outside the old walls of tractability, to make progress on the very goals originally core to semantics.

---

[3] **Goodale and Mascarenhas** acknowledge the need to connect to something beyond language by connecting semantics to symbolic theories of thought. Yet there is no more consensus on the discrete, symbolic, compositional nature of thought than there is for language. To make their point, they favorably cite Quilty-Dunn et al. on Language of Thought (Quilty-Dunn et al., 2023). But Quilty-Dunn et al. discuss probabilistic Language of Thought in depth and approvingly, writing: "We're happy to let a thousand representational formats bloom."

### 4.3.3  Real Patterns Meet Mechanistic Interpretability

So what, then, is the relationship between elegant formal theory and messy implemented systems? We argued that Dennett's (1989) concept of Real Patterns gives us a way to think about linguistic structure as it exists both in human minds and in neural networks. On this view, linguistic theory provides an effective theory of important aspects of observable linguistic behavior—it is valid and real on those terms, even when its relationship to the underlying reductive mechanisms is not clear. That is, linguistic theory remains real even if it does not correspond to a hard-coded innate circuit in a brain or in a neural net.

Two commentaries take up these questions, in particular. **Nefdt and Ladyman** question whether real patterns reside in the language or in the head, and whether we got the Real Patterns account a bit backwards. Drawing on the famous Conway Game of Life analogy, **Nefdt and Ladyman** point out that in the Game of Life, complex patterns (the effective theory) emerge from simple deterministic rules (the reductive theory). But the way we see it, linguistic reality (whether in human brains, language models, or as an abstract system) is messy, complex, and fuzzy—and linguistic theory is a high-level account on top of that.

We don't see our view as contrary to Real Patterns. The Game of Life is underlyingly simple at the reductive level, but other canonical Real Patterns, like beliefs and desires, are notoriously complex at the reductive level. We think beliefs are a better analogy than the Game of Life: there is likely nothing strictly "in the head" that consistently corresponds to a belief—Dennett points out we believe things we've never thought, like "the moon is not made of cheese." Nonetheless, the concept of "belief" is fruitful for prediction. The point is that a pattern can be real—genuinely explanatory, not just a convenient fiction—even when it does not map neatly onto any single mechanism in the system that gives rise to it. We think the formal objects discovered by linguistic theory plausibly work in this way.

These questions raise important topics in mechanistic interpretability, the field that seeks to understand how AI models work mechanistically. **Millière** is sympathetic to the Real Patterns account, but suggests we may actually *undersell* the potential contribution of mechanistic interpretability for understanding exactly how a linguistic task is solved. Our c-command example was intended to illustrate a contrast between what we might hope to find from these methods and what we actually find. The most promising recent methods rely on causal intervention to identify meaningful processes in LMs (Geiger et al., 2025b; Mueller et al., 2026). As linguists or cognitive scientists, we ideally want to know: how is the LM performing complex linguistic tasks? Is it using classical linguistic representations and machinery to represent things like filler–gap dependencies, or has it found a different way to do it?

Because of their distributed nature, it's unlikely we are ever going to see something that looks so neat and tidy that it clearly identifies itself as a classical linguistic algorithm. Instead, we are able to make claims like: by causally intervening on this particular part of the network, we can observe shared structure between related constructions (Boguraev et al., 2025). But these claims require care. It's possible, for instance, to use an overly expressive interpretability model to find structure that isn't really there (Geiger et al., 2025a; Sutter et al., 2025). Nothing is going to replace good, careful science.

But that's not to say that things are hopeless: after all, brains aren't going to reveal neat and tidy algorithms either. Brains are distributed systems, and many of the computations they perform are approximate. If your criterion for 'has linguistic competence' is 'has a circuit that clearly, losslessly, and exactly implements a certain function' then you won't find it in a brain or in a giant neural net. But the task of finding structure in LMs using mechanistic techniques is a useful guide for understanding the relationship between lower-level processes and higher-level theory. Therefore, like **Millière**, we are excited about what mechanistic interpretability will do for questions in linguistics and cognitive science and the ability to "test, in unprecedented detail, which algorithms statistical learners converge on to solve linguistic tasks."

Nevertheless, because of the difficulty in interpreting not only neural networks but of interpreting interpretability results, we are perhaps more skeptical than **Millière** about how clean the answers will be as to which algorithms are the real ones. It's possible that we will have to get more comfortable with shades of gray when it comes to linguistic representation and algorithms. We see this work—figuring out the right methods for interpretability and then deploying them for linguistic and cognitive insights–as one of the most important areas for future work not just in linguistics-focused AI, but in AI research more generally.

## 5 Towards a More Expansive LM Research Program

Many commentators criticize the current LM research program. In many cases, we fully agree with these criticisms. Where we disagree is treating present limitations as principled ceilings—what we call the **As Good As It Gets Assumption**.[4] This assumption proceeds by identifying a limitation of current LMs or of current linguistic research using LMs, and then treats these as fundamental limitations. LMs are still in their infancy, and so is LM research. We see critiques as calls for *more* research into LMs, to see whether these really are limitations. As such, what follows is what we see as a promising "yes, and" research program that acknowledges the limitations and then proposes ways to investigate them. Doing exactly this kind of work is what we need, in order to reveal whether these limitations will point to fundamental ways in which our central thesis is too optimistic.

### 5.1 Learning: data size, data diversity, and acquisition-relevance

We agree with **Ambridge** that "disruption is mutually assured" between LMs and research in language acquisition. Ambridge and a number of other commentaries draw our attention to data used to train models and how it compares (unfavorably in richness, although favorably in number of words) to the input a child gets. **Lampinen** highlights data diversity, and the relative lack of diversity in internet text data compared to the kind of data children learn from. **Wilcox and Newport** call on a need for "human-sized data" and discuss the ways that LMs can help us understand how "graded, fuzzy constraints" interact during language acquisition, leading to a richly realized complex system. **Howitt and Lidz** and particularly **McDermott-Hinman and Feiman** discuss the ways that LM learning trajectories fundamentally differ from those of human learners.

Studying these aspects of how models learn is a promising research program. Indeed, we think efforts, inspired by child language learning, to train smaller models (Warstadt et al., 2023) under various manipulations (Misra and Mahowald, 2024; Patil et al., 2024; Yao et al., 2025) and study learning has been one of the most promising contributions of linguistics to LM research.[5] The commentators lay out promising avenues for further work in these domains. If it turns out that certain linguistic structures are fundamentally not amenable to being learned on human-scale data or that learning trajectories are fundamentally not recoverable by models, that will be an informative data point.

---

[4] An interesting historical example of this fallacy is in Pinker and Prince's 1988 famous argument against Rumelhart and McClelland's 1987 early connectionist work on the English past tense. Pinker and Prince (1988) focus a lot of their argument on criticizing the use of unordered phonetic features called "Wickelphones" as the basic unit in the model. This criticism is valid: it was in hindsight a strange methodological choice to use these kinds of features, a discarded idea unlikely to show up in any of the neural network work being done today. Based on the arguments made and the amount of ink devoted to Wickelphones, one gets the sense that it seemed to Pinker and Prince (1988) and many that it was a fundamental limitation of the whole connectionist program for language. But it wasn't: it was just a particular choice made in that particular implementation. When it came to neural implementations, Wickelphones were not as good as it got.

[5] This line of work on small, custom-trained models is worth bearing in mind when considering **Bolhuis et al.**'s concern about the amount of data and energy needed to train models. They note that Google has ordered "six or seven small nuclear reactors for its AI data centers," but this conflates industrial-scale AI with the models actually used in linguistic research. Many models used in linguistics, including the ones mentioned above, involve training small models from scratch—running in hours on a modest GPU or even a laptop. A back-of-the-envelope calculation: training such a model at home costs around 36 cents in electricity, similar to baking a cake or doing laundry. The nuclear reactors just aren't a useful comparison for many of the models studied in the science of language.

## 5.2 Beyond standard written English: multilinguality, dialects, interaction

Several commentators suggest that, by arguing for the role of LMs in linguistics, we privilege the role of standard written English, particularly syntax and semantics. **Tripp** writes: "LMs as understood in this frame effectively erase ways of languaging which are not readily captured in large repositories of text strings. Multimodal signals, admixtures of codes and the use of sign languages are all ignored, implicitly treated as complexity which is justifiably lost in message 'compression.'"

While much of this is a fair criticism of current LM research, the arguments we make in our target article also apply in principle to future models trained on non-standardized dialects, on audio, on sign language, on admixtures of codes, on multimodal interactive systems. Model building can and should encompass these domains. And we see the body of work on written English not as the endpoint, but as a playbook for future work in all of these areas.

We do agree with **Tripp** that, in our review of the literature, we overfocused on syntax at the expense of other aspects of "languaging". We did so largely both because of the historical importance of syntax in linguistics and cognitive science, and also because that's been the locus of most linguistic research on LMs (probably for the same reason: historical import). We enthusiastically endorse the pleas to push beyond the current boundaries. Below, we outline a few ways we think that can and will happen.

**Non-English Languages** Like **Tripp**, **Ármannsson et al.** point to failures of LMs in other languages, particularly highlighting Icelandic. Modern LM architectures likely favor English (Blasi et al., 2022), and more engineering effort has gone into English-language models. These are genuine shortcomings. We see expanding the scope of languages under study as one of the highest-priority items on the research agenda both as an ethical imperative (since we want systems that work for everyone) and a scientific one (since we want to know how generalizable LM-based results are across languages, and how typological differences interact with inductive biases, model performance, learning trajectories, and the like).

Having said that, we do not think that the current narrowness of LM-based results should *necessarily* be taken to make the extant body of LM work uninformative about language in general. For instance, consider the claim that English LMs deliver proof-of-concept that language *in general* can be learned by general-purpose learners with weak inductive biases. It would be scientifically surprising if English were learnable by large neural systems while typologically different languages were not learnable even in principle (even though other languages might benefit from different architectures or training regimes). A long-held tenet of linguistics is that all languages are learnable by children on roughly similar timescales. Thus, if English were uniquely amenable to statistical approaches, whereas other world languages were impervious to statistical learning, we would find that extremely surprising and thus extremely informative.

**Audio** We agree with **de Heer Kloots et al.** that linguists should also learn to love audio-based LMs. Doing so would extend the insights of this research program into the auditory domain, with import for phonetics and phonology. There is a lot of work left to be done towards this goal. At the same time, while text data lacks the multimodal richness of child input, we think it's sometimes understated how rich it is. While traditional corpora used in corpus linguistics came from news text and books, modern LMs are often trained on almost the whole internet. This includes all kinds of formal writing, but also lots of informal writing: texts, web posts, and so on. In many cases, these data sources approximate informal real-time conversations, using emoji or other cues. So, while extending LM work to audio is an important future goal, we again do not think the extant literature is radically undermined by being text-focused.

**Interaction and Communication** Several commentators pointed out that there is something sterile about LMs: they lack the social, embodied, goal-laden usage that is the hallmark of human language. In that vein, we are sympathetic to much of **Bunzeck et al.'s** argument that LMs find patterns in data but are not fundamentally as "usage-based" as we say since they lack interaction. **Culbertson et al.** make a related point, drawing on the language evolution literature to argue that communication provides a crucial counteracting pressure that prevents language from collapsing into highly predictable but non-functional systems. The extent to which LMs are subject to these pressures is unclear. We find this perspective compelling and agree that the communicative function of language is essential to any complete explanatory account.

There is something off about merely learning from co-occurrence statistics, without having rich communicative goals. And it's true that classical pretrained models fall short in that way. But, as we move past the era of text-based pretrained models (as we largely have), we are optimistic that these limitations can be overcome.

As such, we see the lack of interaction and communication not as a strict limit, but as a demonstration that the right training regime matters. Regimes incorporating communicative objectives— whether through reinforcement learning from human feedback, multi-agent interaction, or grounded language tasks—are the path forward. For instance, with the rise of AI agents, we expect to see a sharp increase in the amount of linguistics-focused work exploring AI behavior in much richer settings. While simple next-word predictors might not have goals or intentions of their own, an AI agent may well inhabit a setting that requires them to have goals (e.g., helping a user write successful code, complete a task). These agents may well have the kind of interactive, communicative setting that some commentators found lacking. Progress in AI is happening so fast that studying linguistic behavior only in non-communicative next-word prediction machines may soon look quaint.

In that vein, we share **Sripada et al.**'s optimism about the ability of LM-based linguistics to make contact with not just how an agent says something, but how it knows what to say when. They revive an old and somewhat dormant idea in linguistics: the Creative Aspect of Language Use (CALU), the ability of language users to select what they will say out of all the possible utterances they could say. We share **Sripada et al.**'s intuition that CALU has lain dormant, in part, because the problem was just too hard for the methods we had. And we agree modern AI models may change that. While there has been a lot of focus on text-based LMs, AI models in general can (and now often do) have visual components, information from various other modalities, and rich representations of the world. They are ready-made for grounding since all of these inputs can exist in a shared vector space, right along with the language. In that sense, "saying the right thing" can become a matter of saying the right thing conditioned on a particular non-linguistic state. We think this plausibly makes much richer modeling of choosing utterances "appropriate to the situation and linguistic context" far more tractable.

# 6    Separate the Science From the Technology

Several commentators (**Resnik**; **Tripp**; **Murphy et al.**; **Dingemanse and Cuskley**) point out the irony of our target article's title, a reference to Kubrick's Atomic Age satire *Dr. Strangelove or: How I Learned to Stop Worrying and Love the Bomb*. In our adaptation, "The Bomb" becomes "the Language Models". Some also noticed that another Kubrick film mentioned in our article, "2001: A Space Odyssey" in the introduction, is also an example where language technology (in the form of HAL) does not go particularly well for humans. We did not arrive at these Kubrick references with eyes wide shut. Indeed, we see the atomic metaphors as apt: the current moment in AI is a major technological moment that will significantly impact humanity in ways we can anticipate and ways we can't. While our take for what we can learn about human language on a scientific level is optimistic, we also see the potential for AI harm. We expect that a lot of human research and effort will be needed to make the AI Age "go well," much as there was massive effort required to make the Atomic Age "go well" and avoid catastrophe.

To that end, we urge language science to stay on the right side of **Dingemanse and Cuskley**'s statement: "Robust research centers values over technology, moving from unexamined technopositivism to a concern with doing the best science possible." We advocate for LM research not out of unexamined techno-positivism, but exactly because we find it scientifically valuable. **Dingemanse and Cuskley** point to a number of shortcomings of existing work with AI, including the closed-source nature of commercial models and lack of reproducibility. The academic community is well-positioned to tackle these challenges and continue to uphold scientific standards.

As such, we would re-focus **Dingemanse and Cuskley**'s argument on how the academic and non-profit communities have risen to the very challenges they lay out. They worry about training data "ever since BERT" being a "carefully guarded and legally fraught secret," but now open-source models like Pythia (from EleutherAI, a non-profit) and OLMo (from the non-profit Allen Institute for AI) have fully transparent training pipelines. "Their frictionless design blinds us to underlying processes and transformations," and yet a thriving field of mechanistic interpretability focuses on understanding their underlying processes. "Their blackboxed nature endangers reproducible research," but efforts within the AI community have focused on replicating and making reproducible workflows. We don't see the project of studying LMs as trading one blackbox (brain) for another, but of trading one blackbox for a system for which we know exactly what it was trained on, how its architecture is structured, and what kind of tools seem most promising for understanding its inner workings.

The project of understanding LMs is part of a broader scientific enterprise, one that has already been fruitful. The field of connectionism developed prototypes that eventually led to models that could produce apparently fluent language. The scientific community took these models apart, probed them, and asked whether they could *really* do what they seemed to. In some cases they couldn't, and that led to improvements. In other cases they could, and the question became how. This is instructive as a model for AI more generally. Just as LMs can inform our understanding of language and cognition, the scientific toolkit from linguistics and cognitive science can contribute to ongoing projects in AI interpretability and safety—projects where a robust, reproducible, reliable science is even more crucial.

We continue to believe that, as a field, linguistics has an important role to play in this enterprise. It is a gift to the field that a key to unlocking today's AI capacities has been the Language Model, an object steeped in linguistic and cognitive relevance. In many ways, we are optimistic that the field is meeting the occasion: a robust and healthy LM-infused science has emerged within many sub-areas of linguistics in the last several years. We hope to continue to learn about human language and human cognition from the most linguistically capable model humanity has ever built.

# 7 Acknowledgments

For helpful comments in preparing this draft, we thank Daniel Drucker. For comments on the draft, we thank Robbie Kubala, Kanishka Misra, and Chris Potts. KM acknowledges funding from NSF CAREER grant 2339729. Generative AI (Claude and ChatGPT) was used for brainstorming, summarizing, and finding relevant literature.